\documentclass[10pt,twocolumn,letterpaper]{article}

\usepackage[pagenumbers]{cvpr} 

\usepackage{graphicx}
\usepackage{amsmath}
\usepackage{amssymb}
\usepackage{booktabs}
\usepackage[linesnumbered,ruled]{algorithm2e}
\usepackage[noend]{algpseudocode}
\usepackage{multirow}
\usepackage{subcaption}
\usepackage{adjustbox}
\usepackage{amsthm}

\newtheorem{proposition}{Proposition}
\usepackage{apxproof}
\usepackage{pifont}
\newcommand{\cmark}{\ding{51}}%
\newcommand{\xmark}{\ding{55}}%

\DeclareMathOperator*{\argmin}{\arg\!\min}
\DeclareMathOperator*{\argmax}{\arg\!\max}
\DeclareMathOperator{\sign}{sign}
\DeclareMathOperator{\E}{\mathbb{E}}
\SetKwInput{KwInput}{Input}                
\SetKwInput{KwOutput}{Output}              

\newcount\Comments  
\usepackage{color}
\newcommand{\kibitz}[2]{\ifnum\Comments=1\textcolor{#1}{#2}\fi}
\newcommand{\ming}[1]{\kibitz{red} {[Ming: #1]}}
\newcommand{\maxma}[1]{\kibitz{blue} {[Max: #1]}}

\usepackage[pagebackref,breaklinks,colorlinks]{hyperref}

\usepackage[capitalize]{cleveref}
\crefname{section}{Sec.}{Secs.}
\Crefname{section}{Section}{Sections}
\Crefname{table}{Table}{Tables}
\crefname{table}{Tab.}{Tabs.}

\def\cvprPaperID{7922} 
\def\confName{CVPR}
\def\confYear{2022}

\begin{document}

\title{Learning Representations Robust to Group Shifts and Adversarial Examples}

\author{%
  Ming-Chang Chiu\\
  Department of Computer Science\\
  University of Southern California\\
  Los Angeles, CA \\
  {\tt\small mingchac@usc.edu}
  \and
  Xuezhe Ma \\
  Information Science Institute \\
  University of Southern California\\
  Los Angeles, CA \\
  {\tt\small xuezhema@isi.edu}
}
\maketitle

\begin{abstract}
   Despite the high performance achieved by deep neural networks on various tasks, extensive studies have demonstrated that small tweaks in the input could fail the model predictions. 
  This issue of deep neural networks has led to a number of methods to improve model robustness, including adversarial training and distributionally robust optimization.
  Though both of these two methods are geared towards learning robust models, they have essentially different motivations: adversarial training attempts to train deep neural networks against perturbations, while distributional robust optimization aims at improving model performance on the most difficult ``uncertain distributions". 
  In this work, we propose an algorithm that combines adversarial training and group distribution robust optimization to improve robust representation learning. Experiments on three image benchmark datasets illustrate that the proposed method achieves superior results on robust metrics without sacrificing much of the standard measures.
\end{abstract}


\section{Introduction}

\begin{figure*}[t]
     \centering
     \begin{subfigure}[b]{0.32\textwidth}
         \centering
         \includegraphics[scale=0.64]{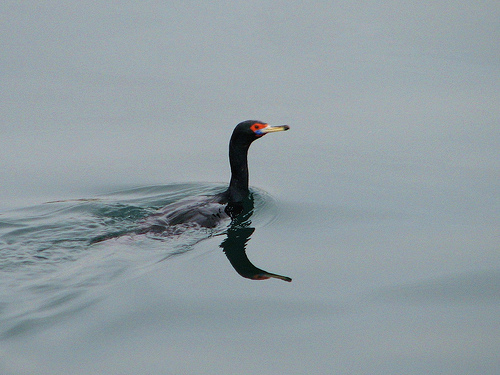}
         \caption{Training example 1 ($y$~: waterbird; $attr$~: water background).}
     \end{subfigure}
     \hfill
     \begin{subfigure}[b]{0.32\textwidth}
         \centering
         \includegraphics[scale=0.36]{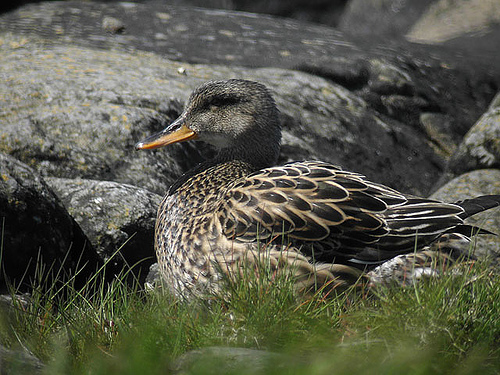}
         \caption{Training example 2 ($y$~: landbird; $attr$~: land background).}
     \end{subfigure}
     \hfill
     \begin{subfigure}[b]{0.32\textwidth}
         \centering
         \includegraphics[scale=0.33,trim=0cm 6cm 0cm 1cm, clip=true]{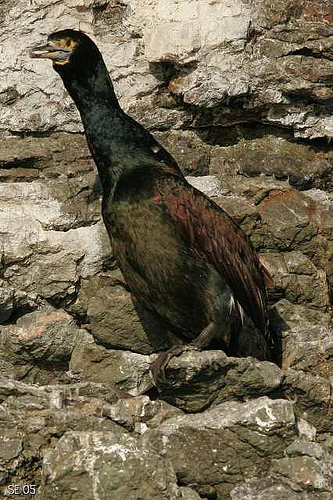}
         \caption{Test example ($y$~: waterbird; $attr$~: land background).}
     \end{subfigure}
        \caption{\textbf{Example of spurious attributes.} Note that the correlations \textit{water} or \textit{land} between bird type $y$ and background $attr$ (short for \textit{attribute}) does not hold at test time.}
        \label{fig:cub_example}
\end{figure*}

Deep neural networks (DNNs) have been demonstrated to significantly improve the benchmark performance in a wide range of application domains, including computer vision \cite{He2016DeepRL}, speech \cite{DBLP:conf/icassp/ChanJLV16}, and natural language processing \cite{Devlin2019BERTPO}.
However, extensive studies have shown that deep neural networks, trained via empirical risk minimization (ERM), are vulnerable: some small and carefully-crafted perturbations in input space can cause malfunctions and huge performance drops~\cite{Goodfellow2015ExplainingAH,DBLP:journals/corr/JiaL17}. 
The essential reason behind performance drop is that the models rely on \textit{weakly correlated} or \textit{spurious correlations}~\cite{DBLP:conf/iclr/TsiprasSETM19}  --- heuristics between labels and inputs that hold for most training examples but are not inherent to the task of interest, such as the strong associations between the background and the label on the Waterbirds dataset~\cite{WahCUB_200_2011} (Figure~\ref{fig:cub_example}).

Adversarial training (AT)~\cite{759851e20d2e47aaad2a560211f6a126,Goodfellow2015ExplainingAH,huang2016learning,Madry2018TowardsDL} is by far one of the most effective ways to learn models against small perturbations \cite{pang2021bag,maini2020adversarial}. The idea behind AT is simple and straight-forward --- adding adversarial noise to the input space during training and therefore achieving better \textit{adversarial robustness} than models trained without AT (\S\ref{bgd}). Previous works have shown various advantages of AT, including mitigating the performance drop on noisy input~\cite{Madry2018TowardsDL,Raghunathan2020UnderstandingAM} or as a regularization technique~\cite{8417973}. 

Another general line of approach toward learning robust models is distributionally robust optimization (DRO)~\cite{10.2307/23359484}. Instead of learning to minimize an ERM objective, DRO aims at \textit{distributional robustness} via optimizing the performance on the worst-case distributions (\S\ref{bgd}). Previous works have proved that DRO is certified to be effective against small perturbations. For example, in \cite{sinha2018certifying}, adversarial robustness is cast as a form of distributional robustness in a Wasserstein ball. 
In this work we will study \textit{group DRO}, which has been shown to reduce the reliance on spurious correlations~\cite{Sagawa2019DistributionallyRN,zhou2021examining}. 
From a view of representation learning, AT and group DRO attempt to improve model robustness of difference aspects: \textit{adversarial robustness} and \textit{group-distributional robustness}, and one straight-forward question is whether we can develop a model to incorporate both of the two types of robustness at the same time. 

In the following sections, we explore the connections between the two types of robustness of AT and group DRO, and propose the \textit{Adversarial group DRO} algorithm, which leverages both the advantages of them to further improve model robustness.
More specifically, we leverage the pre-selected group knowledge in group DRO and the projected gradient descent to learn a group mixture distribution formulated as a minimax problem, robust to group shifts and perturbations.
Experimental results on two datasets with pre-selected spurious features --- Waterbirds~\cite{WahCUB_200_2011,Sagawa2019DistributionallyRN} and CelebA \cite{Liu2015DeepLF} datasets --- demonstrate the effectiveness of the proposed algorithm. 

Our contributions can be summarized in three folds, 
\begin{itemize}
    \item We propose \textit{Adversarial group DRO}, an efficient online optimization algorithm that combines group DRO and AT to improve model robustness. 
    \item Our algorithm shows superior results than simply doing either AT or group DRO, and can mitigate performance drop for robust models on standard dataset like CIFAR-10.
    \item We provide insightful intuitions and supporting evidences on \textit{the learned robust representations} through various types of analysis.
\end{itemize}

\section{Background}
\label{bgd}
In this section, we first set up basic notations and then present the frameworks adopted in this work with brief discussions on their respective issues and connections.

We denote $\mathcal{D}$ as the dataset, and $\langle x, y\rangle$ as a data sample (the image and the corresponding label). $f(\cdot; \theta)$ denotes a deep neural network, which takes an $\langle x, y\rangle$ pair as input.
$\theta$ is the set of parameters of the neural network.
$\mathcal{L}(\cdot, \cdot)$ denotes a generic loss function (e.g., cross-entropy loss).

\subsection{Empirical Risk Minimization}
Typical machine learning algorithms adopt Empirical Risk Minimization (ERM) framework during training, where we learn a model parameterized by $\theta$ minimizing the empirical risk of $\mathcal{L}(\cdot, \cdot)$ under an empirical distribution $\hat{P}$ derived from training data $\mathcal{D}_{train}$:
\begin{equation}\label{eq:1}
    \min_{\theta} \mathbb{E}_{(x,y)\sim \hat{P}}~\mathcal{L}(f(x),y; \theta)
\end{equation}
The underlying assumption is that the training and test set are sampled from the same distribution, i.e. \textit{i.i.d.}, and thus we expect the model to generalize on the test set if it has been optimized during training. 
An issue with ERM is when the model encounters a data distribution that is different from $\hat{P}$ at test time, the performance drops rapidly~\cite{zhang2021dive}. The problem setup where the empirical training distribution $\hat{P}$ is different from test data sampled from some different distribution $\hat{P_T}$ is commonly called \textit{distribution shift}.

\subsection{Distributionally Robust Optimization (DRO)}
To mitigate the issue arising from ERM, a natural solution is to use DRO~\cite{10.2307/23359484}, which instead minimizes the worst expected risk over a family of distributions $\mathcal{Q}$:
\begin{equation}\label{eq:2}
    \min_{\theta \in \Theta}~\Big\{ ~R(\theta) = \sup_{Q\in\mathcal{Q}} \mathbb{E}_{(x,y)\sim Q}~[{\mathcal{L}(f(x),y; \theta)}]~\Big\}
\end{equation}
where $\mathcal{Q}$ is the uncertain set and $R(\theta)$ is the worst-distribution risk. Since $\mathcal{Q}$ encodes all possible distributions at test time, the model is expected to be robust to distributional shifts. A common choice for $\mathcal{Q}$ is a divergence ball around the training distribution which includes a wide range of distributional shifts. 

However, \cite{duchi19} showed that having such divergence ball could result in overly pessimistic models, and a more realistic setting called \textit{group DRO}~\cite{pmlr-v80-hu18a,DBLP:journals/corr/abs-1909-02060,Sagawa2019DistributionallyRN} is adopted in our work. 
Formally, we define the $\mathcal{Q}$ in \textit{group DRO} a coarse-grained mixture models where $P$ is a mixture of $m$ groups containing $P_g$ where $g\in\mathcal{G}=\{1,...,m\}$, and optimize Eq.~\eqref{eq:2} with $\mathcal{Q}=\{\sum^{m}_{g=1} q_gP_g: \sum^{m}_{g=1} q_g=1, q_g\geq0~~ \forall g\in\mathcal{G}\}$. This formulation allows us to learn models that are robust to \textit{group shifts}. Equivalently, since the unique optimal solution of a linear program happens at a vertex~\cite{bertsimas-LPbook}, we can rewrite the inner optimization of Eq.~\eqref{eq:2} as
\begin{equation}\label{eq:g_dro}
    R(\theta) = \max_{g\in\mathcal{G}} \mathbb{E}_{(x,y)\sim P_g}[\mathcal{L}(x,y;\theta)]
\end{equation}
In practice, we leverage prior knowledge on specific tasks or data to define the groups and the corresponding uncertain distributions. For instance, based upon the bird categories and spurious background attribute, we have four groups for Waterbirds --- \{landbird, land; waterbird, land; landbird, water; waterbird, water\}.
\maxma{Add examples such as waterbirds.}

A nice application of \textit{group DRO} is to avoid the reliance on spurious correlation~\cite{Sagawa2019DistributionallyRN,zhou2021examining}, and we hypothesize this can be improved by another robust training method, the adversarial training.

\subsection{Adversarial Training}
\label{AdvTraining}
Different from \textit{group-distributional robustness} in group DRO, AT aims at \textit{adversarial robustness} against adversarial examples by finding the model that minimizes the loss of the maximally perturbed input so that $f(x+\delta) \neq f(x)$:
\begin{equation}\label{eq:at}
 \hat{\theta}_{AT} = \argmin_{\theta} \mathbb{E}_{(x,y)\sim \mathcal{D}}~\left[~{\max_{\delta \in \Delta}  \mathcal{L}(f(x+\delta),y; \theta)}\right],
\end{equation}
where $\delta$ is the perturbation and $\Delta$ is the perturbation distribution. $\Delta$ is designed to be limited in a small boundary to be imperceptible to human eyes~\cite{Goodfellow2015ExplainingAH, Kurakin2017AdversarialML}. For example, given a small budget $\epsilon$, $\Delta:=\{\delta: \|\delta\|_{p}\leq\epsilon\}$ where $\|\cdot\|_{p}$ is the $L_p$ norm. In our work we conduct a number of projected gradient steps to solve for the inner maximization~\cite{Madry2018TowardsDL,Liu2020AdversarialTF} 
\begin{equation}\label{eq:pgd}
    \begin{split}
    g_{adv} \gets \sign(\nabla_{\delta^{(t)}}\mathcal{L}( f(x+\delta^{(t)}),y;\theta)) \\
    \delta^{(t+1)} \gets \Pi_{\|\delta^{(t)}\|\leq\epsilon}(\delta^{(t)} +\eta_{\delta} g_{adv})
    \end{split}
\end{equation}
where $\Pi(\cdot)$ is the projection function, $\eta_{\delta}$ is the adversarial step size, and initial perturbation $\delta^{(0)}$ is sampled from a normal distribution, $\mathcal{N}(0, \sigma^2I)$. We refer readers to \cite{bai2021recent} for a recent survey on AT. 

\begin{algorithm}[t]
\caption{Adversarial group DRO }
\label{alg}
    \KwInput{Step sizes: $\eta_q, \eta_{\theta}, \eta_{\delta}$; $T$: total number of iterations, $\epsilon$: perturbation bound, $\Pi$: projection function, $\sigma^2$: variance of the noise initialization; $K$: the number of perturbation estimation steps, $P_{g}$ for each $g \in G$}
    \For{$t = 1, ..., T$}
    { $g \sim$ Uniform(1,...,m) \tcp*{choose a group}
      $x,y \sim P_g$\tcp*{sample batch}
      $\delta \sim \mathcal{N}(0,\sigma^2 I)$\tcp*{sample noise}
       \For{$k=1,...,K$}{
            $g_{adv} \gets q_g^{(t-1)}\sign(\nabla_{\delta}\mathcal{L}( f(x+\delta),y;\theta^{(t-1)}))$\tcp*{Get gradient direction}
             $\delta \gets \Pi_{\|\delta\|\leq\epsilon}(\delta +\eta_{\delta} g_{adv})$ \tcp*{Ascent step and projection back to $L_p$ ball}
        }
      $q' \gets q^{(t-1)}; q_{g}' \gets q_{g}'exp(\eta_q \mathcal{L}( f(x+\delta),y;\theta^{(t-1)}))$\tcp*{update group weights}
      $q^{(t)} \gets q'/\sum_{g'}q'_{g'}$\tcp*{re-normalize}
      $\theta^{(t)} \gets \theta^{(t-1)} - \eta_{\theta}q_g^{(t)}\nabla_{\theta} \mathcal{L}( f(x+\delta),y;\theta^{(t-1)})$\tcp*{update model}
    }
    \KwOutput{model $\theta$}

\end{algorithm}

\section{Proposed algorithm}
\label{sec_alg}

In this section, we describe the proposed \textit{Adversarial group DRO} algorithm, which effectively combine group DRO and AT to incorporate both adversarial and distributional robustness. Our idea is to train the model under a dynamically changing group mixture distribution where the constituent distributions are adversarially perturbed. This way our model is exposed to both \textit{distributional shifts} (in our case \textit{group shifts}) and \textit{adversarial perturbations}. 

\subsection{Relation between Adversarial Training and DRO}

We emulate Eq.~(\ref{eq:at}) to combine DRO and AT and study the connections of the two types of robustness. We add perturbations into the DRO setup (Eq.~(\ref{eq:2})), and then find the model that optimizes the risk over all the maximally perturbed uncertain distributions:
\begin{multline}\label{eq:adv_dro}
\min_{\theta}~\Big\{ R(\theta) := \\ \sup_{Q\in\mathcal{Q}} \mathbb{E}_{(x,y)\sim Q}\left[{\max_{\delta \in \Delta} \mathcal{L}(f(x+\delta),y; \theta)}\right]~\Big\}
\end{multline}
In our case, DRO carries the group-mixture distributions, so Eq.~(\ref{eq:adv_dro}) can be together considered with Eq.~(\ref{eq:g_dro}) and find the group adversarial model
\begin{multline}\label{eq:adv_gp_dro}
    \theta_{AdvDRO} = \argmin_{\theta}~\Big\{ \\ 
    \max_{g\in\mathcal{G}} \mathbb{E}_{(x,y)\sim P_g}~\left[{\max_{\delta \in \Delta}  \mathcal{L}(f(x+\delta),y; \theta)}\right]~\Big\}.
\end{multline}

\subsection{Adversarial group DRO algorithm}
\label{sec:alg_alg}

Training group DRO and AT jointly can be tricky, as previous works fail for \textit{group DRO} due to the difficulty in gradient estimation in a stochastic fashion~\cite{Duchi2018LearningMW,pmlr-v80-hashimoto18a,Sagawa2019DistributionallyRN} or assume convexity and therefore not generalizable~\cite{sinha2018certifying}.
We propose an online algorithm that provides an efficient way to train Eq.~\eqref{eq:adv_gp_dro}.

Building on top of existing algorithms for \textit{group DRO}~\cite{Sagawa2019DistributionallyRN} and AT~\cite{Liu2020AdversarialTF,Kurakin2017AdversarialML}, Algorithm~\ref{alg} leverages prior knowledge of group information and learns which groups to amass stronger perturbations. Typical AT adds perturbations to the input space uniformly, while our algorithm performs AT phase and optimizes DRO part in turns, which allows us to update the $q$ distribution over groups and weigh perturbations. Essentially, we are learning an adversarial distribution that generates the strongest perturbations to add to each group. 

Note that we can also rewrite Eq.~\eqref{eq:adv_gp_dro} as 
\begin{multline}\label{eq:adv_gp_dro_mixture}
 \hat{\theta}_{AdvDRO} = \argmin_{\theta}~\Big\{ \\
 \max_{q\in\mathcal{Q}} \sum_{g=1}^{m} q_g \mathbb{E}_{(x,y)\sim \hat{P}_g}~\left[{\max_{\delta \in \Delta}  \mathcal{L}(f(x+\delta),y; \theta)}\right]~\Big\},
\end{multline}
where $\mathcal{Q}=\{\sum^{m}_{g=1} q_gP_g: \sum^{m}_{g=1} q_g=1, q_g\geq0~~ \forall g\in\mathcal{G}\}$, so in practice, we can use mini-batches which contain mixture of different groups. And in an end-to-end manner, the algorithm dynamically learns to perform under an ``uncertain distribution" perturbed and mixed with groups and encodes the \textit{group-distributional robustness} and \textit{adversarial robustness}. 

\ming{new} We study error $\epsilon_{T}$ of the average iterate $\Bar{\theta}^{(1:T)}$ and then analyze the convergence rate:
\begin{equation}
    \epsilon_T = \max_{q\in\mathcal{Q}}  L(\Bar{\theta}^{(1:T)},q) - \min_{\theta\in\Theta}\max_{q\in\mathcal{Q}} L(\theta,q),
\end{equation}
where $L(\theta,q):=\sum_{g=1}^m q_g \E_{(x,y)\sim P_g}[\max_{\delta \in \Delta}  \mathcal{L}(f(x+\delta),y; \theta)]$ is the expected worst-case adversarial loss.
Applying Danskin's theorem and results from \cite{nemir09,Sagawa2019DistributionallyRN}, we show in Proposition \ref{alg:pf} that Algorithm \ref{alg} has a standard convergence rate of $O(1/\sqrt{T})$ in a convex setting.

\begin{proposition}
\label{alg:pf}
Suppose that the loss $\mathcal{L}(\cdot; (x, y))$ is non-negative, convex, $B_{\nabla}$-Lipschitz continuous, and bounded by $B_{\mathcal{L}}$ for all $(x, y)$ in $\mathcal{X} \times \mathcal{Y}$, and $\Vert\theta\Vert_{2} \leq B_{\Theta}$ for all $\theta \in \Theta$ with convex $\Theta \subseteq R^{d}$.
Then, the average iterate of Algorithm 1 achieves an expected error at the rate 
    \begin{equation}
        \E[\epsilon_{T}] \leq 2m \sqrt{\frac{10[B^{2}_{\Theta}B^{2}_{\nabla} + B^{2}_{\mathcal{L}}\log m]}{T}}
    \end{equation}
\end{proposition}

\begin{proofsketch}
    We prove \ref{alg:pf} in two steps. We first use Danskin's Theorem to show that the inner-most maximization of \eqref{eq:adv_gp_dro_mixture} is convex and differentiable and then by Proposition 2 of \cite{Sagawa2019DistributionallyRN}, we get its convergence. The full proof is included in the supplemental material.
\end{proofsketch}

\paragraph{A representation learning view.} Although AT and group DRO achieve different kinds of robustness, they both aim to learn robust models. From a representation learning perspective, the model adopting these approaches together should learn correlations that rely less on the spurious ones --- either explicit (as in Figure~\ref{fig:cub_example}) or implicit (innate within the dataset or added by adversarial noise).

\begin{table*}
  \centering
  \begin{adjustbox}{max width=\textwidth}
  \begin{tabular}{llcccc}
    \toprule
    Dataset & Split  & \multicolumn{4}{c}{Subgroup Size ($Y, attr$)} \\
    \midrule
    \multirow{4}{*}{Waterbirds} & & landbird, land & landbird, water & waterbird, land & waterbird, water \\\cline{3-6}
                            & train & 3498 & 184 & 56 & 1057 \\
                            & val & 467 & 466 & 133 & 133\\
                            & test &  2255 & 2255 & 642 & 642\\
                              \hline 
    \multirow{4}{*}{CelebA} & & non-blonde, female & non-blonde, male & blonde, female & blonde, male\\\cline{3-6}
                            & train  & 71629 & 66874 & 22880 & 1387 \\
                            & val & 8535 & 8276 & 2874 & 182 \\
                            & test & 9767 & 7535 & 2480 & 180 \\
    \bottomrule
  \end{tabular}
  \end{adjustbox}
  \caption{We study datasets where spuriously-correlated attribute is present and evaluate the effectiveness of our \textit{adversarial group DRO} algorithm on average and on robust metrics.}
  \label{tab:dataset}
\end{table*}

\begin{table*}
  \centering
  \begin{adjustbox}{max width=\textwidth}
  \begin{tabular}{llcccc|cccc|cccc}
    \toprule
    \multirow{3}{*}{Metric} & \multirow{3}{*}{Perturbation}  & \multicolumn{4}{c}{CIFAR-10}  & \multicolumn{4}{c}{Waterbirds} & \multicolumn{4}{c}{CelebA}\\
        & & \multicolumn{2}{c}{ERM}  & \multicolumn{2}{c}{GDRO} & \multicolumn{2}{c}{ERM}  & \multicolumn{2}{c}{GDRO} & \multicolumn{2}{c}{ERM}  & \multicolumn{2}{c}{GDRO}\\
       & & w/o AT & AT & w/o AT & AT & w/o AT & AT & w/o AT & AT & w/o AT & AT & w/o AT & AT\\
    \midrule
    \multirow{2}{*}{Average Acc.}& Batch & 92.8 & 91.2  & 91.9 & 91.2 & 97.3  & 97.3  & 96.4 & 96.1 & 95.8  & 96  & 94.8 & 95.2\\
                              & Group (ours) & - & - & \textbf{92.3} & \textbf{92} & -      & -      & 96.2 & 96 & -  & -  & 94.8 & 95.2\\
                              \hline 
    \multirow{2}{*}{Adversarial Acc.}& Batch  & 73.4  & 87  & 67.3 & 87.7 & 0  & 38.3  & 0.6 & 32.4 & 73  & 95.1  & 36.1 & 95.3\\
                              & Group (ours)  & -  & -  & \textbf{69.9} & \textbf{88.2} & -   & -    & 0.2 & \textbf{33.6} & -  & - & 29.4 & 93.5\\
                              \hline
    \multirow{2}{*}{Robust Acc.} & Batch  & 87.2 & 84  & 86.5 & 82.8 & 73.5  & 75.2  & 86.2 & 86.2 & 70.5 & 73  & 86.6 & 86.6\\
                              & Group (ours)  & -  & - & 85 & \textbf{86.2} & - & - & 85.8 & \textbf{89.1} & -  & -  & \textbf{90.8} & \textbf{86.6}\\
                              \hline
    
    \multirow{2}{*}{Robust Adv. Acc.} & Batch  & 53.7  & 77  & 53.6 & 78.6 & 2.2  & 55.5  & 17.8 & 60.8 & 1.6  & 37.7  & 5.5 & 83.3\\
                              & Group (ours)  & -  & -  & \textbf{56} & \textbf{79.2} & -  & - & \textbf{17.9} & \textbf{64.5} & -  & -  & \textbf{10.2} & \textbf{83.8}\\
    \bottomrule
  \end{tabular}
  \end{adjustbox}
  \caption{\textbf{Test results (\%).} Our \textit{adversarial group DRO} algorithm improves on the robust metrics on both clean and perturbed test set without sacrificing much of the average accuracy. Note that ERMs do not know group information during training. The difference between Batch and Group is --- in Batch we do not consider group information for adversarial perturbation, we perturb the whole batch uniformly, while in Group we take group weights into account to learn adversarial noises (cf.\S~4.2).}
  \label{tab:acc}
\end{table*}

\begin{table}
  \centering
  \begin{adjustbox}{max width=\linewidth}
  \begin{tabular}{llcccc}
    \toprule
    \multirow{3}{*}{Metric} & \multirow{3}{*}{Perturbation}  & \multicolumn{4}{c}{CIFAR-10} \\
        & & \multicolumn{2}{c}{ERM}  & \multicolumn{2}{c}{GDRO} \\
       & & w/o AT & AT & w/o AT & AT \\
    \midrule
    \multirow{2}{*}{Average Acc.}& Batch & 92.1 & 89.7  & 89.7 & 89.6 \\
                              & Group (ours) & - & - & 89.6 & 89.6\\
                              \hline 
    \multirow{2}{*}{Adversarial Acc.}& Batch  & 22.3 & 79.2  & 31.5 & 77.2 \\
                              & Group (ours)  & -  & -  & \textbf{31.6} & \textbf{77.7} \\
                              \hline
    \multirow{2}{*}{Robust Acc.} & Batch  & 66.1 & 73.8  & 78.5 & 79.3 \\
                              & Group (ours)  & -  & - & \textbf{78.9} & \textbf{81.5} \\
                              \hline
    
    \multirow{2}{*}{Robust Adv. Acc.} & Batch  & 2.5  & 51.4  & 19.4 & 59.8 \\
                              & Group (ours)  & -  & - & 14.8 & \textbf{61.3}\\
    \bottomrule
  \end{tabular}
  \end{adjustbox}
  \caption{\textbf{Additional test results on CIFAR-10.} ($\epsilon=8/255$)}
  \label{tab:acc_cifar10_eps8}
\end{table}

\section{Experiments}
 
\subsection{Experimental Setup}
\paragraph{Datasets.}
To demonstrate the effectiveness of \textit{Adversarial Group DRO} (Algorithm~\ref{alg}), we conduct extensive experiments on three image benchmark datasets and study the connections between \textit{adversarial robustness} and \textit{group-distributional robustness}.

\vspace{2mm}
\noindent \textit{Waterbirds \& CelebA. }  Following~\cite{Sagawa2019DistributionallyRN}, Waterbirds and CelebA both contain four groups (two classes $\times$ two spurious correlations), which are $Y \in$ \{landbird, waterbird\} and $attr \in$ \{land, water\} for Waterbirds, and $Y \in$ \{female, male\} and $attr \in$ \{non-blonde, blond\} for CelebA, with each group having a unbalanced number of examples. Table~\ref{tab:dataset} presents detailed statistics and usages.

\vspace{2mm}
\noindent \textit{CIFAR-10.}  Without manual spurious correlations to form groups, we treat each class of CIFAR-10 as a group in our experiments. Notice that the groups are different from class labels especially for Waterbirds and CelebA, and by nature the groups are different from CIFAR-10 classes since they contain manually crafted spurious features. We hold out 10\% of the training set as validation data.

\paragraph{Implementation details.}
For a single experiment, we use two NVIDIA Tesla P100 GPUs. All our CNN models use SGD as optimizers. Due to resource limits, on CIFAR-10, we train ResNet-110 from scratch with a batch size of 128 and learning rate $\eta_{\theta}=0.1$. On CelebA and Waterbirds, we use pre-trained ResNet-50 with a batch size of 110 and $\eta_{\theta}=0.001$. To train robust models, we perturb input images with max perturbation boundary $\epsilon=2/255$ on a $L_{\infty}$ ball, initial Gaussian noise $\delta^{(0)}$ with $\sigma=\epsilon^2$, and a step size $\eta_{\delta}=0.01$ for 5 steps. Additionally on CIFAR-10, we test Algorithm \ref{alg} with $\epsilon=8/255$. On DROs we set group update rates $\eta_{q}=0.01$. We did not fine-tune hyperparameters extensively and only set them to standard values used by previous works, and we believe fine-tuning can further improve the results. 

\paragraph{Methods.} We train models with objectives described in Section~\ref{bgd} \& \ref{sec_alg}, which are (1) \textit{ERM}, (2) \textit{adversarial ERM} (advERM, i.e., AT), (3) \textit{group DRO} (GDRO), and (4) our algorithm \textit{Adversarial group DRO} (advGDRO). We expect the models to gain \textit{adversarial robustness} from ERM to advERM (so is GDRO to advGDRO) and gain an additional \textit{distributional robustness} from advERM to advGDRO, and thus continuously rely fewer spurious correlations going from ERM to advGDRO. For convenience, we will simply call \textit{group DRO} as DRO in the following sections.

\paragraph{Robust metrics.}
\label{metrics}
We evaluate on average accuracy and average adversarial accuracy to compare \textit{adversarial robustness}. For \textit{distributional robustness}, we use robust accuracy, which can be quantified
by measuring the worst-case performance among all groups. Finally, we measure robust adversarial accuracy for a combined \textit{adversarial} and \textit{distributional robustness}. These metrics are of interest for ERM, AT, GDRO and advGDRO. If we could improve on robust adversarial accuracy, then we essentially improve both types of robustness.

\begin{figure*}[t]
     \centering
     \begin{subfigure}[b]{0.49\textwidth}
         \centering
         \includegraphics[width=\linewidth]{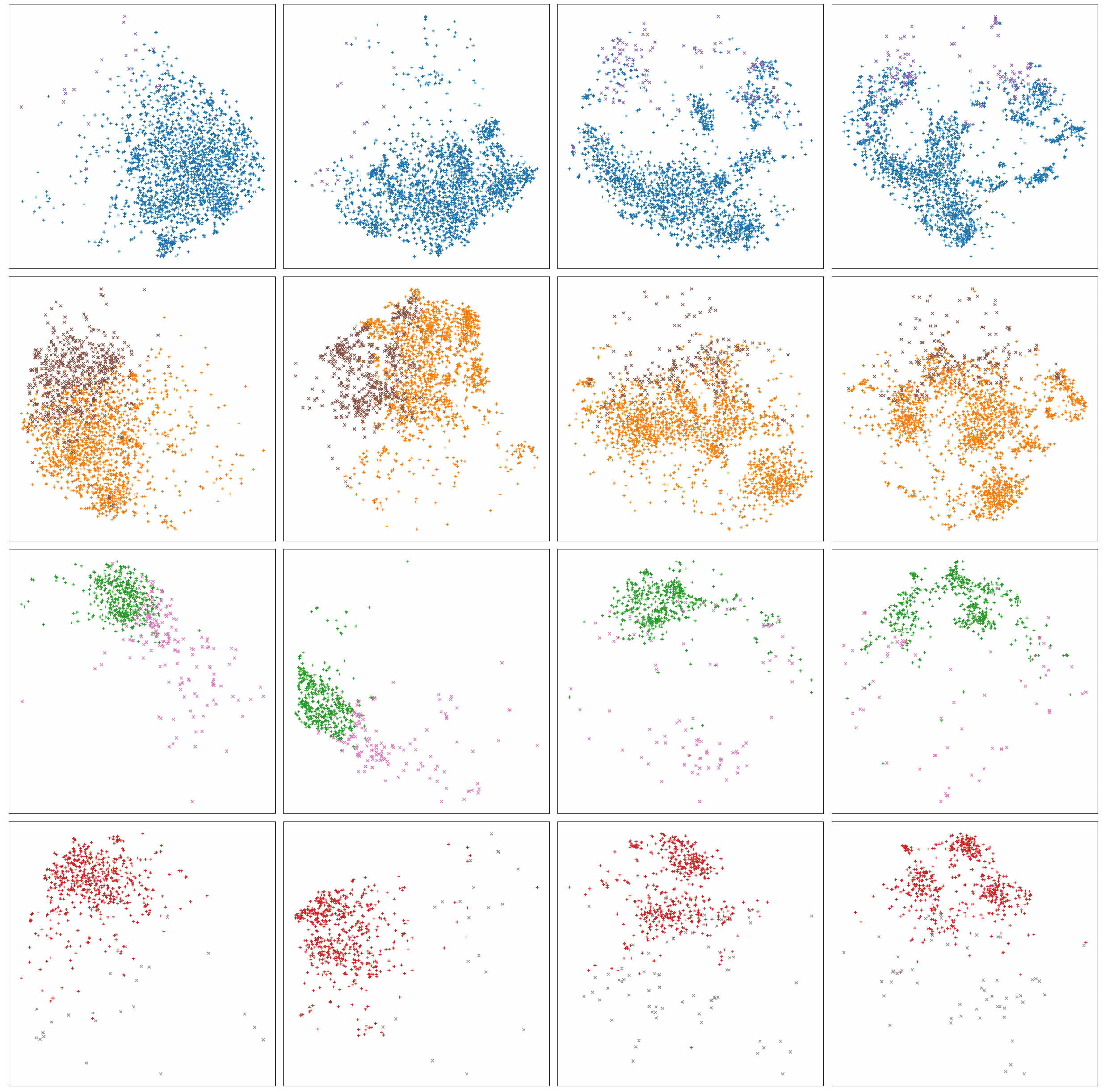}
         \caption{clean test set}
     \end{subfigure}
     \hfill
     \begin{subfigure}[b]{0.49\textwidth}
         \centering
         \includegraphics[width=\linewidth]{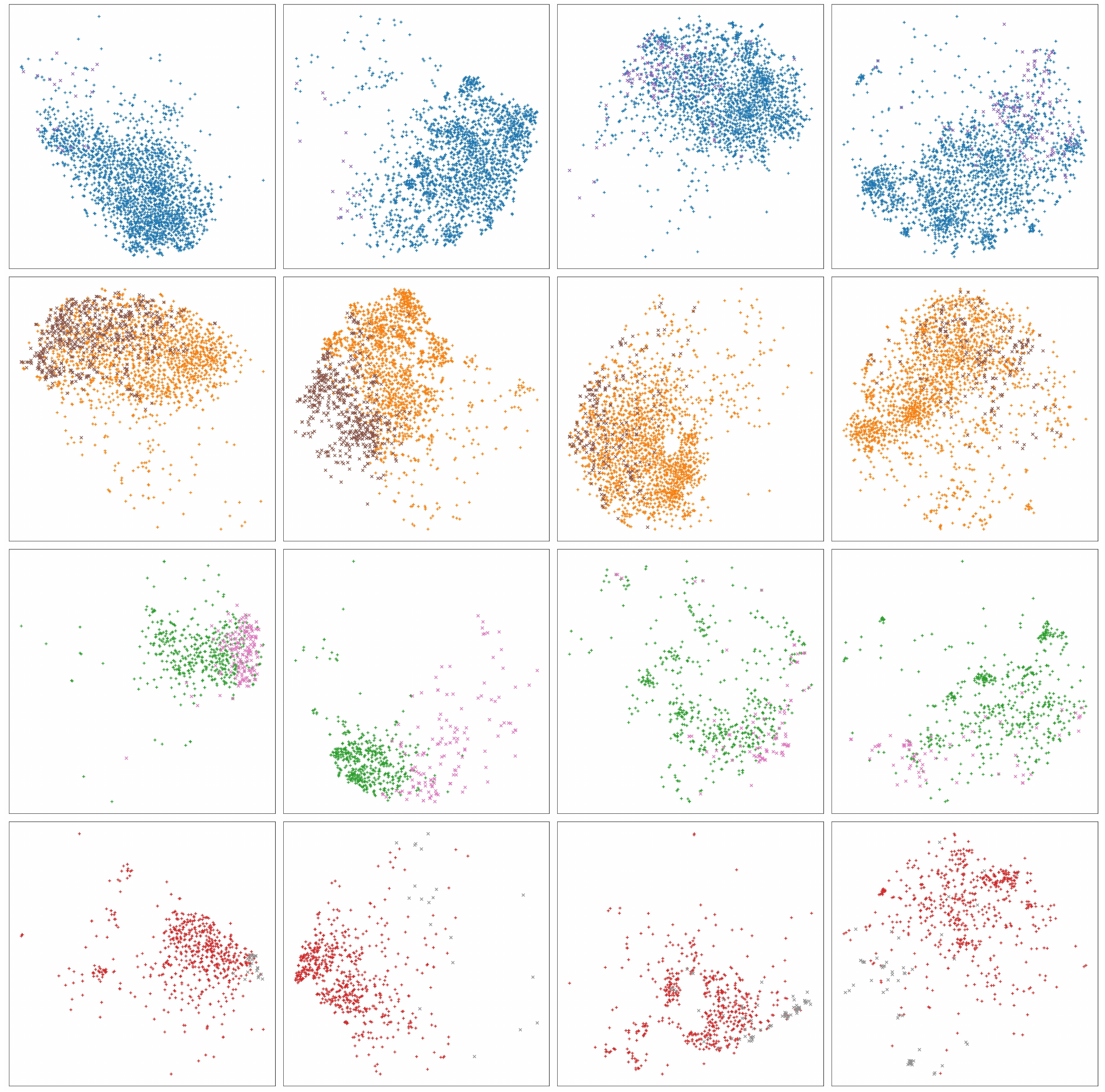}
         \caption{perturbed test set}
     \end{subfigure}
     \caption{\textbf{t-SNE visualizations on Waterbirds.} Output of the last CNN layer (before fc ). For (a)~\&~(b), each row represent a different group; the columns from left to right are (1) ERM, (2) advERM, (3) GDRO, and (4) advGDRO. The majority color stands for the correction predictions, and the minority, wrong predictions. In (a), the data points tend to spread into distinctive clusters as training method becomes more robust, and we believe that our algorithm may help the representations become more disentangled, bringing about better performance. While in (b), the trend is not as obvious but still the data points become more spread out on the plane; we think it is because perturbed data add more spurious correlations, and the robust training has its limit.}
     \label{fig:cub_tsne}
\end{figure*}

\paragraph{Model selection.}
\label{model_selection}
All models are evaluated at the best early stopping epoch as measured by robust metrics on validation set. This way we make sure our results are not overfitting towards the robust metrics.

\subsection{Comparisons and Analysis}
\label{sec_comparison} 

We first compare advGDRO with the three baseline methods, i.e. ERM, AT and GDRO,  on the three benchmark datasets to illustrate the benefits of \textit{adversarial group DRO}.
The experiment results are summarized in Table \ref{tab:acc} \& \ref{tab:acc_cifar10_eps8}. Recall that subgroup information is  available to models only \textit{during training}.

\paragraph{DROs achieves better group-distributional robustness over ERMs and advGDRO further achieves adversarial robustness.} Across the datasets, advGDRO gains at most 13.9\% over advERM on robust accuracy; and GDRO improves over ERM by up to 15.7\% and advDRO gains at most 46.1\% over advERM on robust adversarial accuracy. It shows that advGDRO achieves superior robustness on both clean and perturbed data, and verifies that DROs guarantees better robust performances. The fact that advGDRO consistently outperform other methods on the robust adversarial accuracy demonstrates the effectiveness of our algorithm on improving both types of robustness. 

\paragraph{\textit{Adversarial group DRO} mitigates performance gap.}
Another interest of our work is to mitigate the performance drop that comes with AT \cite{Madry2018TowardsDL} on standard dataset like CIFAR-10. We observe that with mild perturbation \ming{new}our algorithm mitigates the gap on average accuracy from 1.6\% to 0.8\%. In addition, our algorithm surprisingly improve the adversarial accuracy over advERM by 1.2\%, where advERM is designed to optimize against adversarial perturbations.

\paragraph{Group weights incorporated increases adversarial robustness.}
A key benefit of Algorithm~\ref{alg} is to leverage group information to learn the adversarial distribution for Eq.~\eqref{eq:adv_gp_dro_mixture}; however, our algorithm also has the flexibility of perturbing without group weights. To illustrate the effect of group information, we compare the models that are trained with and without group updates. Table \ref{tab:acc} \& \ref{tab:acc_cifar10_eps8} shows improvements on GDRO and the efficacy of using group updates for perturbation is most obvious when combined with advGDRO. When the group information is incorporated, the performance on both worst-group robust measures is consistently improved. On robust accuracy group updates reaches up to 3.4\% performance gain and 3.7\% on robust adversarial accuracy.


                              

\begin{figure*}[t]
     \centering
     \includegraphics[width=\textwidth]{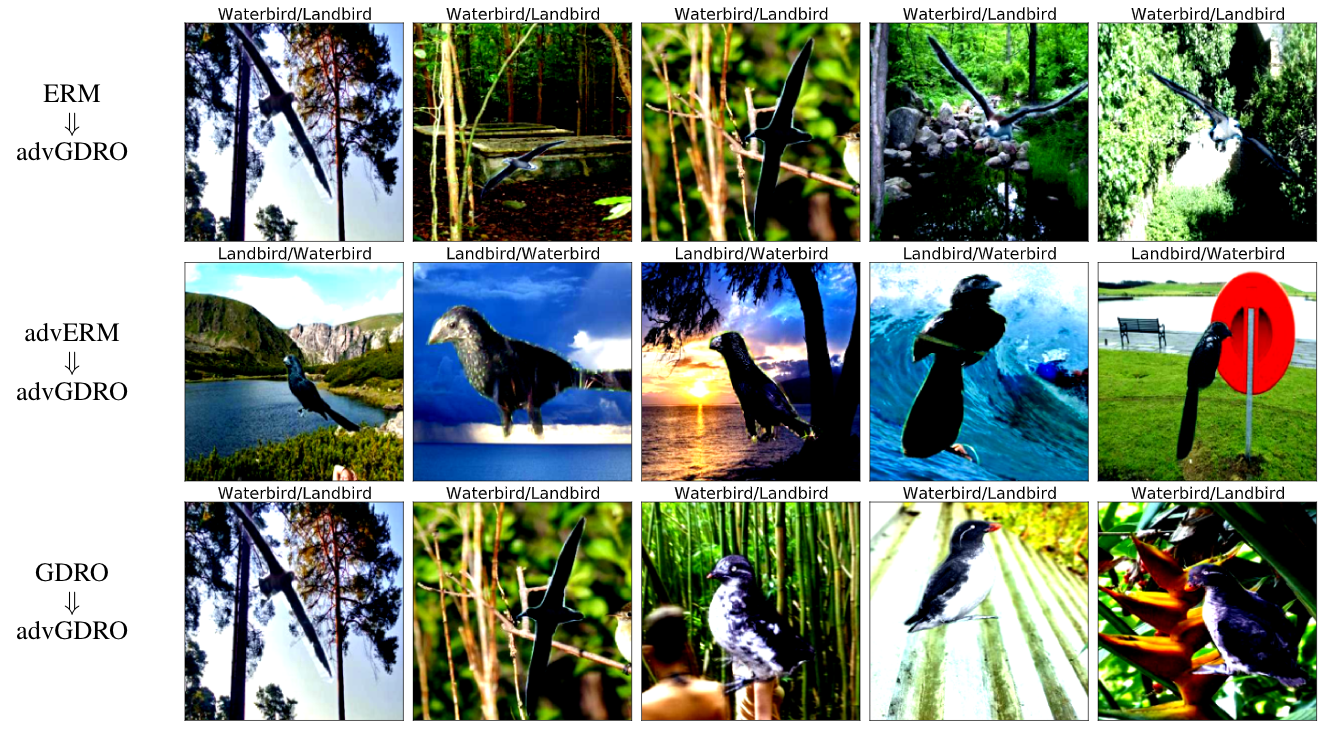}
        \caption{\textbf{advGDRO can correct mis-predictions from all other models.} Row 1 shows the image predictions corrected from ERM to advGDRO; row 2 shows the images corrected from advERM to advGDRO; row 3 shows the images corrected from GDRO to advGDRO. Title of each image is the \textit{prediction of robust model} (\cmark)/\textit{prediction of comparing model} (\xmark).}
        \label{fig_cmp_cub}
\end{figure*}

\subsection{Visualization Analyses}
We discuss the effect of our \textit{Adversarial group DRO} algorithm through the lens of representations and the test examples corrected by more robust models to further analyze what is driving the improvements in this section.
\paragraph{Representation changes show learning to disentangle.} We use \textit{t-SNE}~\cite{JMLR:v9:vandermaaten08a} to visualize the representations of the last ResNet~\cite{He2016DeepRL} layer output (before \verb fc ~layer) in Figure~\ref{fig:cub_tsne}. On clean test set (Figure~\ref{fig:cub_tsne}(a)), we can observe the change of data points distributions from ERM to advGDRO --- over ERMs, the dataset representations have only one cluster; however, going into DROs, each group forms into more disentangled clusters and the disentanglement is most obvious on advGDRO. As indicated in~\cite{scholkopf2021causal}, disentanglement aligns with the goal of \textit{robustness}, i.e., our advGDRO steps toward learning meaningful representations that is robust. On perturbed test set (Figure~\ref{fig:cub_tsne}(b)), though not as obvious as Figure~\ref{fig:cub_tsne}(a), a similar trend can be observed -- the data points get more sparsely scattered as models get more robust. We hypothesize it is because perturbations by nature add noises to images, resulting in more spurious correlations, and thus harder to disentangle; in other words, Figure~\ref{fig:cub_tsne}(b) explains the performance drop and the limit of robust learning.

\paragraph{Corrections from models to models show less reliance on manual spurious correlation.} 
The worst-performing group that the models end up having \textit{a posteriori} also give us some signals for the effect of a spurious attribute; for example, on Waterbirds, the most common worst groups are \{waterbird, land; landbird, water\}, which means the spurious attribute ``background" is a factor that affects the model. Should our method could mitigate this effect, we can correct mistakes made by a less robust model (e.g. ERM). Therefore, we plot out samples that are mistakenly predicted by a less robust model and yet corrected by a robust model. Figure \ref{fig_cmp_cub} shows such samples on the Waterbirds. For example, in row 2, the group was \{landbird, water\} and advERM predicted them as Waterbirds but advGDRO can successfully make the right prediction; in row 1~\&~3, advGDRO can make correct predictions on group \{waterbird, land\}. In other words, advGDRO is the most robust against spurious attributes and can prevent learning them.

\subsection{Computational efficiency} AT is previously known to require longer training time till completion given a total number of epochs; we empirically find that in our setup the run time of our algorithm is only less than 1\% slower than ERM and differs with AT by less than or around 5\%, showing the efficiency of our approach. 

\begin{figure*}[t]
     \centering
     \begin{subfigure}[b]{0.49\textwidth}
         \centering
         \includegraphics[scale=0.20]{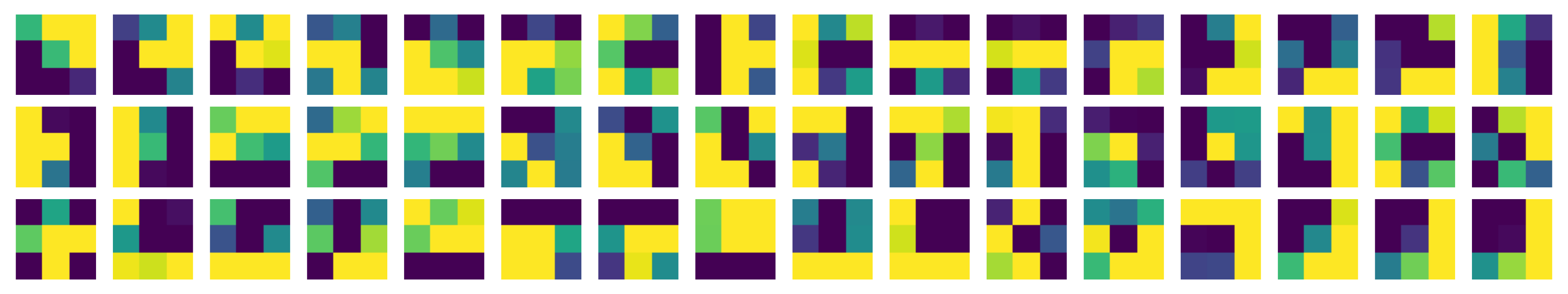}
         \caption{standard ERM}
     \end{subfigure}
     \hfill
     \begin{subfigure}[b]{0.49\textwidth}
         \centering
         \includegraphics[scale=0.20]{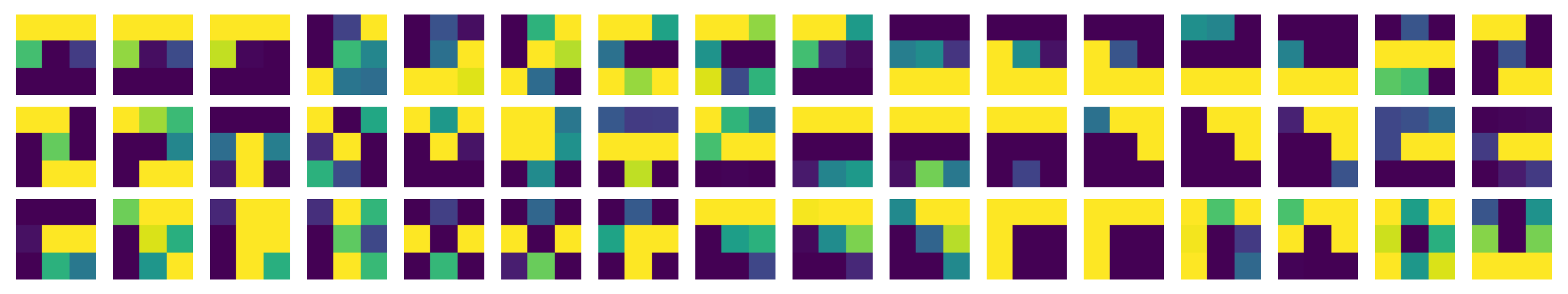}
         \caption{adversarial ERM}
     \end{subfigure}
     \vfill
     \begin{subfigure}[b]{0.49\textwidth}
         \centering
         \includegraphics[scale=0.20]{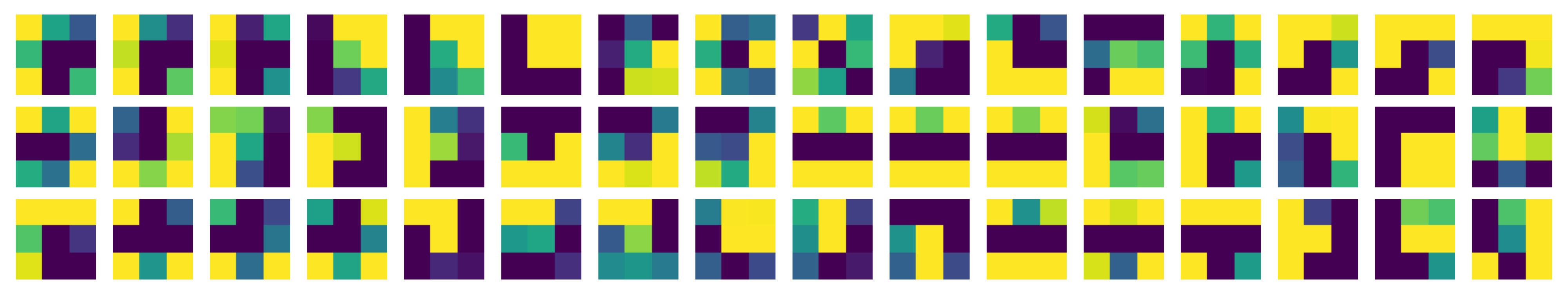}
         \caption{standard DRO}
     \end{subfigure}
     \hfill
     \begin{subfigure}[b]{0.49\textwidth}
         \centering
         \includegraphics[scale=0.20]{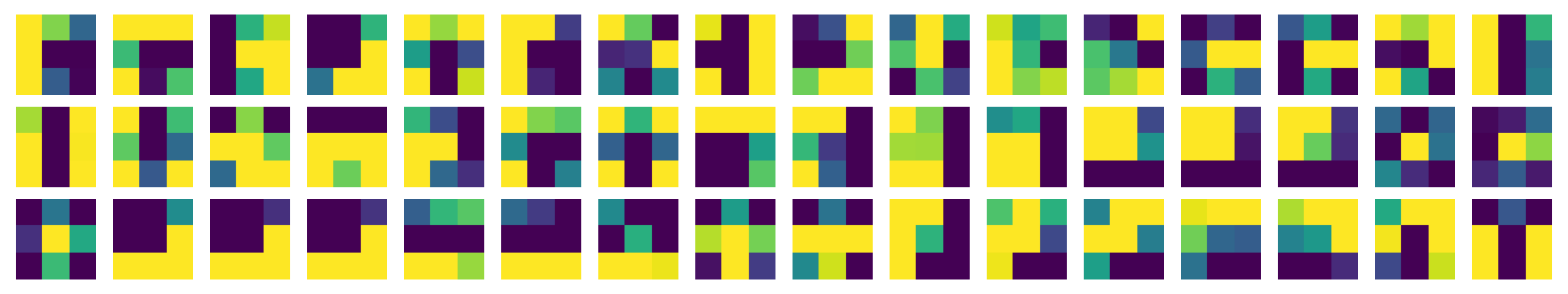}
         \caption{adversarial DRO}
     \end{subfigure}
        \caption{\textbf{Visualization of CNN kernels} (16 kernels each channel $\times$ 3 channels at the first layer). According to \cite{Wang2020HighFrequencyCH}, adversarially robust model should have smoother kernels, and our method (\textit{advGDRO}) produces similar outcome.}
        \label{fig:kernel}
\end{figure*}

\section{Robustness and filter smoothness}
We plot the kernels of the first convolutional layer of CNN on CIFAR-10 and draw connection with~\cite{Wang2020HighFrequencyCH} to further help to see what representations the model learn. \cite{Wang2020HighFrequencyCH} proposes that the filters should be smoother when the model is adversarially trained; and when kernels are regularized to be smoother, the model are stronger against FSGM and PGD. In Figure~\ref{fig:kernel}, we see that the kernels from advGDRO (Figure~\ref{fig:kernel}(d)) are smoother than others, meaning DRO indeed helps adversarial robustness. Notably, the fact that the kernels get smoother from Figure~\ref{fig:kernel}(b) to (d) is also congruent with our observations that DRO help getting the model more robust against adversarial attacks.

\section{Related work}

\paragraph{Representations of neural networks.} The success of a deep learning model generally depends its ability of learning more complex and high quality representations than traditional models~\cite{6472238}. \cite{pmlr-v32-donahue14} showed that intermediate layers of CNN tend to learn simple patterns and high level shapes like lines and corners~\cite{Krizhevsky2012ImageNetCW}. These features are essential to the performance of CNNs~\cite{He2015DelvingDI}. Our work attempts to learn robust representations under an adversarial and uncertain setting. 

\paragraph{Attacking DNNs.} An active area of studying DNN behaviors is attacking the DNNs. Researchers have found that DNNs are susceptible to various types of \textit{threat models}, which deceive the models and cause undesired behaviors of DNNs. One type is the backdoor attack --- implanting malicious data into the training data~\cite{DBLP:journals/corr/abs-1708-06733,Liu2018TrojaningAO}, the model learns wrong behaviors and then consistently make the wrong decisions at test time. Another one is the adversarial attack --- ever since~\cite{Goodfellow2015ExplainingAH}, adversarial examples have been broadly studied and a wide range of attacks such as FGSMs and PGD~\cite{Goodfellow2015ExplainingAH,Kurakin2017AdversarialML,45818,Madry2018TowardsDL} have been proposed, and they give rise to a broader discussions of vulnerabilities of neural networks in computer vision~\cite{DBLP:journals/corr/abs-2007-07677,8601309,DBLP:journals/corr/abs-1712-05526,ijcai2018-543,chen2018ead}, speech processing \cite{szurley2019perceptual} and NLP tasks~\cite{DBLP:journals/corr/JiaL17}. 
Meanwhile, understanding adversarial examples~\cite{NEURIPS2019_e2c420d9,Zhao2021WhatDD} has been studied as well. 

\paragraph{Toward robustness.} In order for the models to defend against the deceptions or \textit{threats}, researchers have set out to work on closing the gap between adversarial accuracy and the standard accuracy~\cite{Madry2018TowardsDL,Carlini2017TowardsET}, and a wide range of \textit{defense methods} for different types of attack~\cite{chen2021depois,DBLP:journals/corr/abs-2003-01908,NEURIPS2018_358f9e7b}. Adversarial training, discussed in Section~\ref{AdvTraining}, is the most popular method against adversarial attacks. Previous works demonstrated its capability of working with other frameworks~\cite{Raghunathan2020UnderstandingAM,Xie_2020_CVPR,8417973,NEURIPS2020_5de8a360}, such as self-supervised learning, etc. Discussions about the trade-off between the robustness and generalization have been attempted~\cite{Wang2020HighFrequencyCH,DBLP:journals/corr/abs-1901-08573}. However, a universal method to fully prevent all the aforementioned attacks from happening seems yet to come~\cite{DBLP:journals/corr/abs-1902-06705}.

\paragraph{Robust optimization.} The community has also started to study robustness from an optimization point of view by proving certifiable bounds of the attacks~\cite{DBLP:journals/corr/abs-1906-06316,pmlr-v97-cohen19c}. 
DRO has drawn attention due to its nature to upper-bound the expected risk under an unknown test distribution~\cite{duchi2020learning,10.2307/23359484}, and how the distributions are formed --- either coarse-grained group~\cite{pmlr-v80-hu18a,DBLP:journals/corr/abs-1909-02060,Sagawa2019DistributionallyRN} as we adopted in this work, or other types~\cite{10.2307/23359484,sinha2018certifying,RePEc:inm:oropre:v:58:y:2010:i:3:p:595-612}. How to solve DRO problems under different setups~\cite{ShafieezadehAbadeh2015DistributionallyRL,NIPS2016_4588e674,NIPS2017_5a142a55} have also been proposed and studied. 

\section{Conclusion}
In this paper, we explored the connections between \textit{group-distributional robustness} and \textit{adversarial robustness} and proposed an algorithm for robust representation learning. Using only assumptions about the smoothness of the loss function $\mathcal{L}$, we show that our algorithm enjoys a theoretical guarantee\ming{new}. By achieving improved performances via our algorithm over benchmark datasets, we have made a step toward that goal. Our results show utilizing group weighting end-to-end to learn the ``uncertain distribution", we could further enhance the robustness.

On the representation side, when models are trained robustly, we observed that the representations learned show \textit{disentanglement} on the 2D t-SNE embedding space, and therefore more robust and meaningful. 

The mispredicted images corrected by our algorithm also demonstrate that the proposed algorithm can prevent models from learning pre-defined spurious correlations.

In sum, our work provides a connection for future studies in the robustness of distribution shifts and adversarial training, and on a broader level, the pursuit of learning robust representations.

{\small
\bibliographystyle{ieee_fullname}
\bibliography{references}

\begin{thebibliography}{}

\end{thebibliography}


\begin{thebibliography}{10}\itemsep=-1pt

\bibitem{bai2021recent}
Tao Bai, Jinqi Luo, Jun Zhao, Bihan Wen, and Qian Wang.
\newblock Recent advances in adversarial training for adversarial robustness,
  2021.

\bibitem{10.2307/23359484}
Aharon Ben-Tal, Dick den Hertog, Anja~De Waegenaere, Bertrand Melenberg, and
  Gijs Rennen.
\newblock Robust solutions of optimization problems affected by uncertain
  probabilities.
\newblock {\em Management Science}, 59(2):341--357, 2013.

\bibitem{6472238}
Yoshua Bengio, Aaron Courville, and Pascal Vincent.
\newblock Representation learning: A review and new perspectives.
\newblock {\em IEEE Transactions on Pattern Analysis and Machine Intelligence},
  35(8):1798--1828, 2013.

\bibitem{bertsimas-LPbook}
D. Bertsimas and J.N. Tsitsiklis.
\newblock {\em Introduction to linear optimization}.
\newblock Athena Scientific, 1997.

\bibitem{DBLP:journals/corr/abs-1902-06705}
Nicholas Carlini, Anish Athalye, Nicolas Papernot, Wieland Brendel, Jonas
  Rauber, Dimitris Tsipras, Ian~J. Goodfellow, Aleksander Madry, and Alexey
  Kurakin.
\newblock On evaluating adversarial robustness.
\newblock {\em ArXiv}, abs/1902.06705, 2019.

\bibitem{Carlini2017TowardsET}
Nicholas Carlini and David~A. Wagner.
\newblock Towards evaluating the robustness of neural networks.
\newblock {\em 2017 IEEE Symposium on Security and Privacy (SP)}, pages 39--57,
  2017.

\bibitem{DBLP:conf/icassp/ChanJLV16}
William Chan, Navdeep Jaitly, Quoc~V. Le, and Oriol Vinyals.
\newblock Listen, attend and spell: {A} neural network for large vocabulary
  conversational speech recognition.
\newblock In {\em 2016 {IEEE} International Conference on Acoustics, Speech and
  Signal Processing, {ICASSP} 2016, Shanghai, China, March 20-25, 2016}, pages
  4960--4964. {IEEE}, 2016.

\bibitem{chen2021depois}
Jian Chen, Xuxin Zhang, Rui Zhang, Chen Wang, and Ling Liu.
\newblock De-pois: An attack-agnostic defense against data poisoning attacks.
\newblock In {\em IEEE Transactions on Information Forensics and Security},
  2021.

\bibitem{chen2018ead}
Pin-Yu Chen, Yash Sharma, Huan Zhang, Jinfeng Yi, and Cho-Jui Hsieh.
\newblock Ead: Elastic-net attacks to deep neural networks via adversarial
  examples.
\newblock In {\em AAAI}, 2018.

\bibitem{DBLP:journals/corr/abs-1712-05526}
Xinyun Chen, Chang Liu, Bo Li, Kimberly Lu, and Dawn Song.
\newblock Targeted backdoor attacks on deep learning systems using data
  poisoning.
\newblock {\em ArXiv}, abs/1712.05526, 2017.

\bibitem{pmlr-v97-cohen19c}
Jeremy Cohen, Elan Rosenfeld, and Zico Kolter.
\newblock Certified adversarial robustness via randomized smoothing.
\newblock In {\em Proceedings of the 36th International Conference on Machine
  Learning}, 2019.

\bibitem{RePEc:inm:oropre:v:58:y:2010:i:3:p:595-612}
Erick Delage and Yinyu Ye.
\newblock Distributionally robust optimization under moment uncertainty with
  application to data-driven problems.
\newblock {\em Operations Research}, 58(3):595--612, 2010.

\bibitem{Devlin2019BERTPO}
J. Devlin, Ming-Wei Chang, Kenton Lee, and Kristina Toutanova.
\newblock Bert: Pre-training of deep bidirectional transformers for language
  understanding.
\newblock In {\em NAACL}, 2019.

\bibitem{pmlr-v32-donahue14}
Jeff Donahue, Yangqing Jia, Oriol Vinyals, Judy Hoffman, Ning Zhang, Eric
  Tzeng, and Trevor Darrell.
\newblock Decaf: A deep convolutional activation feature for generic visual
  recognition.
\newblock In Eric~P. Xing and Tony Jebara, editors, {\em Proceedings of the
  31st International Conference on Machine Learning}, volume~32 of {\em
  Proceedings of Machine Learning Research}, pages 647--655, Bejing, China,
  22--24 Jun 2014. PMLR.

\bibitem{NEURIPS2020_5de8a360}
Yinpeng Dong, Zhijie Deng, Tianyu Pang, Jun Zhu, and Hang Su.
\newblock Adversarial distributional training for robust deep learning.
\newblock In {\em Advances in Neural Information Processing Systems}, 2020.

\bibitem{duchi2020learning}
John Duchi and Hongseok Namkoong.
\newblock Learning models with uniform performance via distributionally robust
  optimization.
\newblock {\em ArXiv}, 2020.

\bibitem{Duchi2018LearningMW}
John~C. Duchi and Hongseok Namkoong.
\newblock Learning models with uniform performance via distributionally robust
  optimization.
\newblock {\em ArXiv}, abs/1810.08750, 2018.

\bibitem{Goodfellow2015ExplainingAH}
I. Goodfellow, Jonathon Shlens, and Christian Szegedy.
\newblock Explaining and harnessing adversarial examples.
\newblock In {\em ICLR}, volume abs/1412.6572, 2015.

\bibitem{DBLP:journals/corr/abs-1708-06733}
Tianyu Gu, Brendan Dolan{-}Gavitt, and Siddharth Garg.
\newblock Badnets: Identifying vulnerabilities in the machine learning model
  supply chain.
\newblock {\em ArXiv}, abs/1708.06733, 2017.

\bibitem{pmlr-v80-hashimoto18a}
Tatsunori Hashimoto, Megha Srivastava, Hongseok Namkoong, and Percy Liang.
\newblock Fairness without demographics in repeated loss minimization.
\newblock In {\em ICML}, 2018.

\bibitem{He2015DelvingDI}
Kaiming He, X. Zhang, Shaoqing Ren, and Jian Sun.
\newblock Delving deep into rectifiers: Surpassing human-level performance on
  imagenet classification.
\newblock {\em 2015 IEEE International Conference on Computer Vision (ICCV)},
  pages 1026--1034, 2015.

\bibitem{He2016DeepRL}
Kaiming He, X. Zhang, Shaoqing Ren, and Jian Sun.
\newblock Deep residual learning for image recognition.
\newblock {\em CVPR}, pages 770--778, 2016.

\bibitem{pmlr-v80-hu18a}
Weihua Hu, Gang Niu, Issei Sato, and Masashi Sugiyama.
\newblock Does distributionally robust supervised learning give robust
  classifiers?
\newblock In {\em Proceedings of the 35th International Conference on Machine
  Learning}, 2018.

\bibitem{huang2016learning}
Ruitong Huang, Bing Xu, Dale Schuurmans, and Csaba Szepesvari.
\newblock Learning with a strong adversary, 2016.

\bibitem{NEURIPS2019_e2c420d9}
Andrew Ilyas, Shibani Santurkar, Dimitris Tsipras, Logan Engstrom, Brandon
  Tran, and Aleksander Madry.
\newblock Adversarial examples are not bugs, they are features.
\newblock In {\em Advances in Neural Information Processing Systems}, 2019.

\bibitem{DBLP:journals/corr/JiaL17}
Robin Jia and Percy Liang.
\newblock Adversarial examples for evaluating reading comprehension systems.
\newblock In {\em EMNLP}, volume abs/1707.07328, 2017.

\bibitem{duchi19}
Hongseok~Namkoong John~Duchi, Tatsunori~Hashimoto.
\newblock Distributionally robust losses against mixture covariate shifts,
  2019.

\bibitem{Krizhevsky2012ImageNetCW}
A. Krizhevsky, Ilya Sutskever, and Geoffrey~E. Hinton.
\newblock Imagenet classification with deep convolutional neural networks.
\newblock {\em Communications of the ACM}, 60:84 -- 90, 2012.

\bibitem{45818}
Alexey Kurakin, Ian Goodfellow, and Samy Bengio.
\newblock Adversarial examples in the physical world.
\newblock {\em ICLR Workshop}, 2017.

\bibitem{Kurakin2017AdversarialML}
Alexey Kurakin, I. Goodfellow, and S. Bengio.
\newblock Adversarial machine learning at scale.
\newblock {\em ArXiv}, abs/1611.01236, 2017.

\bibitem{Liu2020AdversarialTF}
X. Liu, Hao Cheng, Pengcheng He, Weizhu Chen, Yu Wang, Hoifung Poon, and
  Jianfeng Gao.
\newblock Adversarial training for large neural language models.
\newblock {\em ArXiv}, abs/2004.08994, 2020.

\bibitem{Liu2018TrojaningAO}
Yingqi Liu, Shiqing Ma, Yousra Aafer, W. Lee, Juan Zhai, Weihang Wang, and X.
  Zhang.
\newblock Trojaning attack on neural networks.
\newblock In {\em NDSS}, 2018.

\bibitem{Liu2015DeepLF}
Z. Liu, Ping Luo, Xiaogang Wang, and X. Tang.
\newblock Deep learning face attributes in the wild.
\newblock {\em ICCV}, pages 3730--3738, 2015.

\bibitem{Madry2018TowardsDL}
A. Madry, Aleksandar Makelov, Ludwig Schmidt, D. Tsipras, and Adrian Vladu.
\newblock Towards deep learning models resistant to adversarial attacks.
\newblock {\em ArXiv}, abs/1706.06083, 2018.

\bibitem{maini2020adversarial}
Pratyush Maini, Eric Wong, and J.~Zico Kolter.
\newblock Adversarial robustness against the union of multiple perturbation
  models.
\newblock In {\em International Conference on Machine Learning}, 2020.

\bibitem{8417973}
Takeru Miyato, Shin-Ichi Maeda, Masanori Koyama, and Shin Ishii.
\newblock Virtual adversarial training: A regularization method for supervised
  and semi-supervised learning.
\newblock {\em IEEE Transactions on Pattern Analysis and Machine Intelligence},
  41(8):1979--1993, 2019.

\bibitem{NIPS2016_4588e674}
Hongseok Namkoong and John~C Duchi.
\newblock Stochastic gradient methods for distributionally robust optimization
  with f-divergences.
\newblock In {\em Advances in Neural Information Processing Systems}, 2016.

\bibitem{NIPS2017_5a142a55}
Hongseok Namkoong and John~C Duchi.
\newblock Variance-based regularization with convex objectives.
\newblock In I. Guyon, U.~V. Luxburg, S. Bengio, H. Wallach, R. Fergus, S.
  Vishwanathan, and R. Garnett, editors, {\em Advances in Neural Information
  Processing Systems}, volume~30. Curran Associates, Inc., 2017.

\bibitem{nemir09}
Arkadi Nemirovski, Anatoli Juditsky, Guanghui Lan, and And Shapiro.
\newblock Robust stochastic approximation approach to stochastic programming.
\newblock {\em Society for Industrial and Applied Mathematics}, 19:1574--1609,
  01 2009.

\bibitem{DBLP:journals/corr/abs-1909-02060}
Yonatan Oren, Shiori Sagawa, Tatsunori~B. Hashimoto, and Percy Liang.
\newblock Distributionally robust language modeling.
\newblock In {\em EMNLP}, volume abs/1909.02060, 2019.

\bibitem{pang2021bag}
Tianyu Pang, Xiao Yang, Yinpeng Dong, Hang Su, and Jun Zhu.
\newblock Bag of tricks for adversarial training, 2021.

\bibitem{Raghunathan2020UnderstandingAM}
Aditi Raghunathan, Sang~Michael Xie, Fanny Yang, John~C. Duchi, and Percy
  Liang.
\newblock Understanding and mitigating the tradeoff between robustness and
  accuracy.
\newblock {\em ArXiv}, abs/2002.10716, 2020.

\bibitem{DBLP:journals/corr/abs-2007-07677}
Jonas Rauber and Matthias Bethge.
\newblock Fast differentiable clipping-aware normalization and rescaling.
\newblock {\em ArXiv}, abs/2007.07677, 2020.

\bibitem{Sagawa2019DistributionallyRN}
Shiori Sagawa, Pang~Wei Koh, T. Hashimoto, and Percy Liang.
\newblock Distributionally robust neural networks for group shifts: On the
  importance of regularization for worst-case generalization.
\newblock {\em ArXiv}, abs/1911.08731, 2019.

\bibitem{DBLP:journals/corr/abs-2003-01908}
Hadi Salman, Mingjie Sun, Greg Yang, Ashish Kapoor, and J.~Zico Kolter.
\newblock Black-box smoothing: {A} provable defense for pretrained classifiers.
\newblock In {\em NeurIPS}, volume abs/2003.01908, 2020.

\bibitem{scholkopf2021causal}
Bernhard Schölkopf, Francesco Locatello, Stefan Bauer, Nan~Rosemary Ke, Nal
  Kalchbrenner, Anirudh Goyal, and Yoshua Bengio.
\newblock Towards causal representation learning, 2021.

\bibitem{ShafieezadehAbadeh2015DistributionallyRL}
Soroosh Shafieezadeh-Abadeh, Peyman~Mohajerin Esfahani, and D. Kuhn.
\newblock Distributionally robust logistic regression.
\newblock In {\em NIPS}, 2015.

\bibitem{sinha2018certifying}
Aman Sinha, Hongseok Namkoong, Aman Sinha, and John Duchi.
\newblock Certifying some distributional robustness with principled adversarial
  training.
\newblock In {\em International Conference on Learning Representations}, 2018.

\bibitem{8601309}
Jiawei Su, Danilo~Vasconcellos Vargas, and Kouichi Sakurai.
\newblock One pixel attack for fooling deep neural networks.
\newblock {\em IEEE Transactions on Evolutionary Computation}, 23(5):828--841,
  2019.

\bibitem{759851e20d2e47aaad2a560211f6a126}
Christian Szegedy, Wojciech Zaremba, Ilya Sutskever, Joan Bruna, Dumitru Erhan,
  Ian Goodfellow, and Rob Fergus.
\newblock Jan. 2014.
\newblock 2nd International Conference on Learning Representations, ICLR 2014 ;
  Conference date: 14-04-2014 Through 16-04-2014.

\bibitem{szurley2019perceptual}
Joseph Szurley and J.~Zico Kolter.
\newblock Perceptual based adversarial audio attacks.
\newblock {\em ArXiv}, 2019.

\bibitem{DBLP:conf/iclr/TsiprasSETM19}
Dimitris Tsipras, Shibani Santurkar, Logan Engstrom, Alexander Turner, and
  Aleksander Madry.
\newblock Robustness may be at odds with accuracy.
\newblock In {\em ICLR}, 2019.

\bibitem{JMLR:v9:vandermaaten08a}
Laurens van~der Maaten and Geoffrey Hinton.
\newblock Visualizing data using t-sne.
\newblock {\em Journal of Machine Learning Research}, 9(86):2579--2605, 2008.

\bibitem{WahCUB_200_2011}
C. Wah, S. Branson, P. Welinder, P. Perona, and S. Belongie.
\newblock {The Caltech-UCSD Birds-200-2011 Dataset}.
\newblock Technical Report CNS-TR-2011-001, California Institute of Technology,
  2011.

\bibitem{Wang2020HighFrequencyCH}
Haohan Wang, Xindi Wu, Pengcheng Yin, and E. Xing.
\newblock High-frequency component helps explain the generalization of
  convolutional neural networks.
\newblock {\em 2020 IEEE/CVF Conference on Computer Vision and Pattern
  Recognition (CVPR)}, pages 8681--8691, 2020.

\bibitem{NEURIPS2018_358f9e7b}
Eric Wong, Frank Schmidt, Jan~Hendrik Metzen, and J.~Zico Kolter.
\newblock Scaling provable adversarial defenses.
\newblock In {\em Advances in Neural Information Processing Systems}, 2018.

\bibitem{ijcai2018-543}
Chaowei Xiao, Bo Li, Jun yan Zhu, Warren He, Mingyan Liu, and Dawn Song.
\newblock Generating adversarial examples with adversarial networks.
\newblock In {\em Proceedings of the Twenty-Seventh International Joint
  Conference on Artificial Intelligence, {IJCAI-18}}, pages 3905--3911.
  International Joint Conferences on Artificial Intelligence Organization, 7
  2018.

\bibitem{Xie_2020_CVPR}
Cihang Xie, Mingxing Tan, Boqing Gong, Jiang Wang, Alan~L. Yuille, and Quoc~V.
  Le.
\newblock Adversarial examples improve image recognition.
\newblock In {\em CVPR}, June 2020.

\bibitem{zhang2021dive}
Aston Zhang, Zachary~C. Lipton, Mu Li, and Alexander~J. Smola.
\newblock Dive into deep learning.
\newblock {\em arXiv preprint arXiv:2106.11342}, 2021.

\bibitem{DBLP:journals/corr/abs-1906-06316}
Huan Zhang, Hongge Chen, Chaowei Xiao, Bo Li, Duane~S. Boning, and Cho{-}Jui
  Hsieh.
\newblock Towards stable and efficient training of verifiably robust neural
  networks.
\newblock {\em ArXiv}, abs/1906.06316, 2019.

\bibitem{DBLP:journals/corr/abs-1901-08573}
Hongyang Zhang, Yaodong Yu, Jiantao Jiao, Eric~P. Xing, Laurent~El Ghaoui, and
  Michael~I. Jordan.
\newblock Theoretically principled trade-off between robustness and accuracy.
\newblock In {\em ICML}, 2019.

\bibitem{Zhao2021WhatDD}
Shihao Zhao, Xingjun Ma, Yisen Wang, J. Bailey, Bo Li, and Yu-Gang Jiang.
\newblock What do deep nets learn? class-wise patterns revealed in the input
  space.
\newblock {\em ArXiv}, abs/2101.06898, 2021.

\bibitem{zhou2021examining}
Chunting Zhou, Xuezhe Ma, Paul Michel, and Graham Neubig.
\newblock Examining and combating spurious features under distribution shift.
\newblock In {\em Proceedings of the 38th International Conference on Machine
  Learning}, 2021.

\end{thebibliography}
}

\newpage
\appendix
\section{Proof of Proposition \ref{alg:pf}}

\noindent
\textbf{Proposition \ref{alg:pf}.} Suppose that the loss $\mathcal{L}(\cdot; (x, y))$ is non-negative, convex, $B_{\nabla}$-Lipschitz continuous, and bounded by $B_{\mathcal{L}}$ for all $(x, y)$ in $\mathcal{X} \times \mathcal{Y}$, and $\Vert\theta\Vert_{2} \leq B_{\Theta}$ for all $\theta \in \Theta$ with convex $\Theta \subseteq R^{d}$.
Then, the average iterate of Algorithm 1 achieves an expected error at the rate 
    \begin{equation}
        \E[\epsilon_{T}] \leq 2m \sqrt{\frac{10[B^{2}_{\Theta}B^{2}_{\nabla} + B^{2}_{\mathcal{L}}\log m]}{T}}
    \end{equation}

\begin{proof} 
We prove Proposition \ref{alg:pf} in two parts. First we prove that the inner-most maximization of \eqref{eq:adv_gp_dro_mixture} is convex and differentiable, and then by Proposition 2 of \cite{Sagawa2019DistributionallyRN}, we get the convergence guarantee. 

Let $F(\theta) := \max_{\delta\in\Delta}\mathcal{L}(f(x+\delta),y, \theta)$. 
Since $\Delta$ is a compact convex set, by Danskin's theorem, if $Z_{0}(\theta):=\{\delta \in \argmax_{\delta\in\Delta}\mathcal{L}(f(x+\delta),y, \theta)\}$ is singleton for some $\theta$, then $F(\theta)$ is convex and directionally differentiable. By Corollary C.2 of \cite{Madry2018TowardsDL}, $F(\theta)$ has an ascent direction, and in practice, we observe most of the elements of $\delta$ reach the boundary after the projected gradient steps. 

Now \eqref{eq:adv_gp_dro_mixture} can be written as a saddle-point problem,
\begin{equation}
    \min_{\theta\in\Theta} \max_{q\in\mathcal{Q}} \sum_{g=1}^{m} q_g \mathbb{E}_{(x,y)\sim \hat{P}_g}~[F(\theta)].
\end{equation}
By Proposition 2 of \cite{Sagawa2019DistributionallyRN}, we can use the result of \cite{nemir09} Eq.(3.23) to obtain a similar bound,
\begin{equation}
    \E[\epsilon_{T}] \leq 2m \sqrt{\frac{10[B^{2}_{\Theta}B^{2}_{\nabla} + B^{2}_{\mathcal{L}}\log m]}{T}}.
\end{equation}

\end{proof} 
\end{document}